\documentclass[twoside]{article}
\usepackage[preprint]{aistats2027}
\usepackage[utf8]{inputenc}
\usepackage[T1]{fontenc}
\usepackage{amsmath,amssymb}
\usepackage[round]{natbib}
\usepackage{url}
\newcommand{\toprule}{\hline}
\newcommand{\midrule}{\hline}
\newcommand{\bottomrule}{\hline}
\newtheorem{theorem}{Theorem}
\newtheorem{proposition}[theorem]{Proposition}
\newtheorem{corollary}[theorem]{Corollary}
\newtheorem{lemma}[theorem]{Lemma}
\newtheorem{definition}[theorem]{Definition}
\newenvironment{proof}[1][Proof]{\par\noindent\textit{#1.}\ }{\hfill$\square$\par}
\newcommand{\R}{\mathbb R}
\newcommand{\E}{\mathbb E}
\newcommand{\Var}{\operatorname{Var}}

\newcommand{\tr}{\operatorname{tr}}
\newcommand{\cond}{\operatorname{cond}}
\newcommand{\argmin}{\operatorname*{arg\,min}}
\newcommand{\norm}[1]{\lVert#1\rVert}
\newcommand{\B}{\mathcal B}
\newcommand{\cP}{\mathcal P}
\newcommand{\front}{\mathfrak b}
\begin{document}
\runningtitle{The Price of Local Updates in Federated Optimization}
\twocolumn[
\aistatstitle{Faster to the Wrong Answer: The Price of Local Updates in Federated Optimization}
\aistatsauthor{Molly Wang}
\aistatsaddress{Imperial Business School}
]

\begin{abstract}
Local computation can improve the conditioning of federated updates but may also introduce bias, causing their fixed point to differ from the global minimizer. We characterize the minimum worst-case bias achievable under a prescribed conditioning bound. We consider strongly convex quadratic client objectives with Hessian eigenvalues in \([\mu,L]\) and local updates represented by a shared positive spectral filter. Client heterogeneity is measured by the root-mean-square gradient at the global minimizer. For a uniform surrogate condition number bounded by \(\chi\in[1,L/\mu]\), we establish the exact worst-case bias per unit heterogeneity, \((L-\mu\chi)/(2\mu L\sqrt{\chi})\), and show that local gradient descent with sufficiently small step sizes and proximal updates attain this bound. Our analysis reduces populations of arbitrary dimension, including noncommuting Hessians, to distributions over scalar curvatures. This reduction provides optimality certificates and worst-case constructions involving at most three scalar clients. We further extend the sharp proximal bias bound to smooth strongly convex objectives beyond quadratics and quantify the error introduced by inexact local solves.
\end{abstract}

\section{Introduction}
\label{sec:intro}
Local computation reduces communication in federated optimization, but faster convergence of the resulting update need not imply convergence to the desired solution. In FedAvg, clients start from a common model, take several steps on their own objectives, and return models that the server averages \citep{mcmahan2017}. On heterogeneous quadratic objectives, these local steps can improve the conditioning of the aggregate update while shifting its fixed point. Quantifying both effects is necessary to assess the accuracy associated with this improvement in conditioning.

Prior work establishes the trade-off but leaves an exact worst-case comparison under a common gradient-heterogeneity budget unresolved. \citet{charles2021} show that quadratic local updates optimize a surrogate objective and compare convergence--accuracy curves. SCAFFOLD and FedSplit correct local directions or change the update structure to recover the intended optimizer \citep{karimireddy2020,pathak2020}. We ask a narrower question: among a specified class of local updates, what fixed-point error is unavoidable when the update must satisfy a prescribed conditioning bound?

We study shared spectral filters because they describe the aggregate effect of repeated local steps on quadratic objectives. Client $i$ has objective $f_i$, Hessian $H_i$, and aggregation weight $p_i>0$, with $\sum_i p_i=1$. Assume the eigenvalues of every $H_i$ lie in $[\mu,L]$, where $0<\mu<L$. A continuous positive function $q$ scales a gradient component with curvature $h$ by $q(h)$. The resulting direction and its uniform surrogate condition number are
\begin{equation}
 \begin{aligned}
 D_q(x)&=\sum_i p_i q(H_i)\nabla f_i(x),\\
 \chi(q)&=\frac{\max_{h\in[\mu,L]} hq(h)}
 {\min_{h\in[\mu,L]} hq(h)}.
 \end{aligned}
 \label{eq:direction-intro}
\end{equation}
Repeated local gradient steps with common parameters and exact proximal solves both take this form. The surrogate condition number bounds the conditioning of the quadratic objective with gradient $D_q$.

The client gradients at the global minimizer measure the disagreement that can shift the fixed point. Let $x_*$ minimize $F=\sum_i p_i f_i$. Define the root-mean-square (RMS) gradient magnitude by $\zeta_*^2=\sum_i p_i\norm{\nabla f_i(x_*)}^2$. Although the weighted gradients cancel at $x_*$, their individual values can be nonzero. Applying different curvature-dependent scalings can destroy that cancellation. The quantity $\zeta_*$ measures this disagreement without requiring gradient differences to remain bounded at every point.

Our main result gives the exact minimum worst-case error at each admissible conditioning bound. Let $x_q$ denote the zero of $D_q$. Then
\begin{equation}
 \begin{aligned}
 \inf_{q:\,\chi(q)\leq\chi}\ \sup_{\zeta_*\leq 1}
 \norm{x_q-x_*}
 &=\front(\chi):=\frac{L-\mu\chi}{2\mu L\sqrt\chi},\\
 &\hspace{-5em}1\leq\chi\leq L/\mu.
 \end{aligned}
 \label{eq:frontier-intro}
\end{equation}
The supremum ranges over all quadratic client populations with the stated curvature and heterogeneity bounds. The constraint $\zeta_*\leq1$ fixes the scale; the error bound grows linearly with any other heterogeneity budget. Bias and conditioning are separate worst-case properties of the update rule, so different populations may attain their respective bounds.

\paragraph{Contributions.}
Three results characterize worst-case bias for the stated local-update class. First, we reduce the worst-case matrix problem exactly to a distribution over scalar curvatures and derive an optimal-design certificate with at most three support points (Section~\ref{sec:reduction}). Second, we prove the frontier in \eqref{eq:frontier-intro} and show that stable local gradient descent and proximal variants attain it (Section~\ref{sec:frontier}). Third, we extend the sharp proximal bound to nonquadratic objectives and quantify local-solve and aggregation residuals (Section~\ref{sec:nonquadratic}). Each part includes explicit instances that attain the claimed constants.

The frontier compares the accuracy and conditioning of existing update rules. It excludes local computation costs and algorithms with control variates or additional persistent state. Two filters can therefore share the same guarantee but require different amounts of work; the frontier does not establish oracle or communication-complexity optimality.

\subsection{Related work}
The relevant literature distinguishes fixed-point bias from convergence rates and drift correction. The closest bias analysis is the surrogate framework of \citet{charles2021}. Their Theorem~4 controls optimizer displacement through bounds on client minimizers; we bound the RMS client gradient at the global optimum. Because gradients also depend on client curvature, these budgets define different admissible populations. Our results give an exact scalar reduction, a frontier over positive filters, and a nonlinear proximal bound under the latter budget. The surrogate interpretation and qualitative trade-off are prior results. We do not address their bounded-minimizer conjecture.

Convergence analyses of local methods address additional questions about stochasticity and computational resources. \citet{woodworth2018,woodworth2020} study communication lower bounds and comparisons with minibatch methods. \citet{glasgow2022} analyze stochastic iterate bias, and \citet{patel2025} obtain bounds that exploit second-order heterogeneity. \citet{mangold2025} derive constant-step FedAvg bias expansions and use Richardson--Romberg extrapolation. Our results instead evaluate the exact deterministic fixed-point error at fixed filter parameters, without assuming Hessian similarity.

Proximal regularization and drift correction provide two distinct ways to change local updates. FedProx adds a penalty for moving away from the server model \citep{li2020}; the corresponding quadratic local solves belong to our filter class. SCAFFOLD and FedSplit introduce corrections that preserve the global optimizer \citep{karimireddy2020,pathak2020}, placing them outside that class.

Our scalar reduction uses classical tools to solve the resulting moment problem. The Kiefer--Wolfowitz condition certifies optimal designs for general filters \citep{kiefer1960}; the Bhatia--Davis variance bound yields the endpoint formula \citep{bhatia2000}.

\section{Client objectives, local updates, and comparison criteria}
\label{sec:setup}
Quadratic objectives let us express the aggregate local update exactly. This class includes regularized least-squares objectives and allows different client curvature matrices and minimizers. We use this model to determine how those differences change the update's fixed point.

\subsection{Objective class and heterogeneity}
The objective class permits arbitrary client weights and Hessians within a common curvature interval. Fix $0<\mu<L<\infty$. A population contains finitely many clients with weights $p_i>0$, $\sum_i p_i=1$, and objectives
\begin{equation}
 \begin{aligned}
 f_i(x)&=\tfrac12 x^\top H_i x-b_i^\top x,\\
 \mu I&\preceq H_i\preceq LI,\qquad
 F(x)=\sum_i p_i f_i(x).
 \end{aligned}
 \label{eq:quadratics}
\end{equation}
The weights represent the server's aggregation rule, for example weighting clients by their sample counts. The Hessians may have different eigenvectors. We allow the dimension and client count to vary when taking worst cases. Throughout the paper, vector norms are Euclidean and matrix norms are spectral. Strong convexity gives $F$ a unique minimizer $x_*$.

Heterogeneity at $x_*$ measures client-gradient disagreement when the aggregate gradient vanishes. Define
\begin{equation}
 g_i=\nabla f_i(x_*),\quad \sum_i p_i g_i=0,
 \quad \zeta_*^2=\sum_i p_i\norm{g_i}^2.
 \label{eq:heterogeneity}
\end{equation}
A positive $\zeta_*$ means that some clients would move away from $x_*$ if they optimized their objectives alone. When $\zeta_*=0$, every client gradient vanishes at the global minimizer.

\subsection{Spectral updates and their two performance criteria}
A spectral filter expresses how local computation scales each curvature component of a client's gradient. For a continuous positive $q$, define $q(H_i)$ by applying $q$ to the eigenvalues of $H_i$ and retaining its eigenvectors. Set $r(h)=hq(h)$ and
\begin{equation}
 \begin{aligned}
 Q_i&=q(H_i),& \overline Q&=\sum_i p_iQ_i,\\
 A_q&=\sum_i p_iH_iQ_i,\\
 m_q&=\min_h r(h),& M_q&=\max_h r(h).
 \end{aligned}
 \label{eq:matrices}
\end{equation}
All minima and maxima range over $[\mu,L]$. Each $H_iQ_i$ is symmetric positive definite, so $D_q$ is the gradient of a strongly convex quadratic with Hessian $A_q$. We call this quadratic the surrogate objective and denote its unique minimizer by $x_q$.

The surrogate minimizer differs from $x_*$ precisely when the filtered client gradients fail to cancel. Substituting the quadratic gradients into $D_q(x_q)=0$ yields
\begin{equation}
 x_q-x_*=-A_q^{-1}\sum_i p_iQ_i g_i.
 \label{eq:bias-identity}
\end{equation}
This is the standard quadratic surrogate identity \citep{charles2021}. Appendix~\ref{app:identity} includes the derivation and the exact worst-case error for fixed Hessians and weights.

We compare filters by their largest possible fixed-point error and surrogate condition number over this objective class.
\begin{definition}[Worst-case bias]
\leavevmode\par\noindent
For the fixed interval $[\mu,L]$, define
\begin{equation}
 \B(q)=\sup\bigl\{\norm{x_q-x_*}:\text{\eqref{eq:quadratics} holds and }\zeta_*\leq1\bigr\}.
 \label{eq:bias-definition}
\end{equation}
The supremum ranges over all dimensions, client counts, positive weights, and vectors $b_i$. Define the uniform condition number by $\chi(q)=M_q/m_q$.
\end{definition}

The conditioning bound describes the entire class rather than a particular population. Every population satisfies $\cond(A_q)\leq\chi(q)$, and one client in dimension two attains equality by placing its eigenvalues at a minimizer and a maximizer of $r$. This population need not maximize the bias. Thus a scalar worst-case bias example is consistent with a uniform condition number greater than one.

The normalization separates message scaling from heterogeneity scaling. Multiplying $q$ by a positive constant leaves $\B(q)$ and $\chi(q)$ unchanged. Scaling the heterogeneity budget gives $\norm{x_q-x_*}\leq\zeta_*\B(q)$, including the case $\zeta_*=0$. These properties let us compare the shape of different filters without fixing an arbitrary message normalization.

\section{Reducing worst-case bias to scalar optimal design}
\label{sec:reduction}
The main obstacle to an exact bias bound is that client Hessians need not share eigenvectors. We remove this obstacle by reducing the matrix problem to a scalar distribution over curvatures. We then use optimal-design theory to certify the maximizing distribution and bound the number of clients needed to attain it.

\subsection{An exact scalar characterization}
One-dimensional clients suffice to attain the worst-case bias, even when the original class allows noncommuting Hessians in arbitrary dimensions. Let $\cP([\mu,L])$ denote the probability measures on $[\mu,L]$, and let $h$ denote the curvature variable. The next theorem expresses the answer through the variance of $q(h)$ and the mean of $r(h)$.

\begin{theorem}[Scalar reduction]
\label{thm:scalar}
\leavevmode\par\noindent
For every continuous positive $q$,
\begin{equation}
 \B(q)^2=\max_{P\in\cP([\mu,L])}
 \frac{\Var_P(q(h))}{(\E_P[hq(h)])^2}.
 \label{eq:scalar}
\end{equation}
A population of at most three one-dimensional clients attains the maximum. Moreover, $\B(q)=0$ if and only if $q$ is constant on $[\mu,L]$.
\end{theorem}

\begin{proof}
We bound every matrix population through a scalar distribution over its curvatures. Fix a unit vector $v$. Centering with the mean-zero condition in \eqref{eq:heterogeneity} and applying Cauchy--Schwarz yields
\begin{equation}
 \left|v^\top\sum_i p_i Q_i g_i\right|^2
 \leq\zeta_*^2\left\{\sum_i p_i\norm{Q_i v}^2-\norm{\overline Qv}^2\right\}.
 \label{eq:centered-bound}
\end{equation}
To express this bound through scalar curvatures, write $H_i=\sum_j h_{ij}u_{ij}u_{ij}^\top$ in an orthonormal eigenbasis. Define $P_v=\sum_{i,j}p_i(u_{ij}^\top v)^2\delta_{h_{ij}}$, where $\delta_h$ is a unit mass at $h$. This is a probability measure, and
\begin{align}
 \sum_i p_i\norm{Q_i v}^2-\norm{\overline Qv}^2
 &\leq\E_{P_v}q(h)^2-(\E_{P_v}q(h))^2,\label{eq:variance-reduction}\\
 \norm{A_qv}&\geq v^\top A_qv=\E_{P_v}r(h).\label{eq:denominator-reduction}
\end{align}
Set $b=\sum_i p_iQ_i g_i$. The dual identity $\norm{A_q^{-1}b}=\sup_{v\ne0}|v^\top b|/\norm{A_qv}$ now proves the upper bound in \eqref{eq:scalar}.

Scalar populations attain the same ratio. For a finite distribution $P$ with masses $p_i$ at curvatures $h_i$ and nonzero filter variance, set $g_i=(q(h_i)-\E_Pq)/\sqrt{\Var_P(q)}$ and $f_i(x)=h_ix^2/2+g_ix$. These choices give $x_*=0$, $\zeta_*=1$, and equality in the ratio. Finite measures approximate arbitrary measures because the integrands are continuous and $r$ stays positive. Proposition~\ref{prop:design} establishes attainment with at most three support points. Finally, constant filters give zero variance, while a two-point distribution gives positive variance whenever $q$ is nonconstant.
\end{proof}

\subsection{Optimality certificates and finite support}
The scalar formula becomes a classical optimal-design problem after reweighting the curvature distribution. This representation is useful for finding hard client populations: it supplies a condition that certifies global optimality rather than only a candidate lower bound. A design measure specifies curvatures and weights; its $2\times2$ information matrix records three moments of the filter.

\begin{proposition}[Design characterization]
\label{prop:design}
\leavevmode\par\noindent
Define $z(h)=r(h)^{-1/2}(1,q(h))^\top$ and $M(\nu)=\E_\nu[z(h)z(h)^\top]$ for a probability measure $\nu$. Then
\begin{equation}
 \B(q)^2=\max_{\nu\in\cP([\mu,L])}\det M(\nu).
 \label{eq:design}
\end{equation}
For nonconstant $q$, a measure $\nu_*$ is optimal if and only if $M_*=M(\nu_*)$ is positive definite and
\begin{equation}
 z(h)^\top M_*^{-1}z(h)\leq2\quad\text{for every }h\in[\mu,L].
 \label{eq:certificate}
\end{equation}
An optimal measure exists with at most three support points. If it assigns mass $w_i$ to $h_i$, the corresponding client weights are proportional to $w_i/r(h_i)$.
\end{proposition}

Reweighting by surrogate curvature converts the variance ratio into a determinant. Specifically, define $d\nu=r\,dP/\E_Pr$. Then
\begin{equation}
 \det M(\nu)=
 \frac{\E_Pq^2-(\E_Pq)^2}{(\E_Pr)^2}.
 \label{eq:tilt}
\end{equation}
The inverse transformation is $dP=(r^{-1}d\nu)/\E_\nu r^{-1}$, so the two optimization problems are equivalent. Maximizing this determinant is the two-parameter $D$-optimal design problem. The Kiefer--Wolfowitz condition gives \eqref{eq:certificate} \citep{kiefer1960}; Appendix~\ref{app:design} proves the condition and derives the support bound from the geometry of the information matrices.

General filters can require more than two curvatures to attain their largest bias. Appendix~\ref{app:three} gives a finite spectral set on which a three-client population has strictly larger bias than every two-client population. For the local gradient and proximal filters considered next, an additional inequality reduces the optimum to the two endpoints $\mu$ and $L$.

\section{The exact frontier and the local methods that attain it}
\label{sec:frontier}
We now minimize worst-case bias subject to a bound on uniform surrogate conditioning. Two endpoint curvatures provide a lower bound for every positive filter. A matching upper bound shows that proximal updates and stable finite local gradient steps attain the resulting frontier.

\subsection{When endpoint curvatures attain the worst case}
An affine lower bound on $r(h)$ reduces the scalar optimization to a two-client problem. Suppose $q$ decreases from $u=q(\mu)$ to $v=q(L)$ with $u>v$. Define the chord connecting the endpoint pairs $(u,r(\mu))$ and $(v,r(L))$:
\begin{equation}
 \ell(s)=\frac{s-v}{u-v}r(\mu)+\frac{u-s}{u-v}r(L).
 \label{eq:chord}
\end{equation}
The next lemma applies whenever the filter values stay between $v$ and $u$ and the curve $(q(h),r(h))$ lies above this chord.

\begin{lemma}[Endpoint certificate]
\label{lem:endpoint}
If $q([\mu,L])\subseteq[v,u]$ and $r(h)\geq\ell(q(h))$ for every $h$, then
\begin{equation}
 \B(q)=\frac{q(\mu)-q(L)}{2\sqrt{\mu Lq(\mu)q(L)}}.
 \label{eq:endpoint-bias}
\end{equation}
The maximizing scalar population has curvatures $(\mu,L)$ and weights $p_\mu=r(L)/(r(\mu)+r(L))$, $p_L=1-p_\mu$.
\end{lemma}

\begin{proof}
The mean filter value controls both terms in the scalar ratio. Write $a=\E_Pq=\theta u+(1-\theta)v$ with $\theta\in[0,1]$. Averaging $(u-q)(q-v)\geq0$ gives $\Var_Pq\leq(u-a)(a-v)=\theta(1-\theta)(u-v)^2$, the Bhatia--Davis inequality \citep{bhatia2000}. The chord assumption gives $\E_Pr\geq\ell(a)=\theta r(\mu)+(1-\theta)r(L)$. Hence the square root of the ratio in \eqref{eq:scalar} is at most
\begin{equation}
 \frac{(u-v)\sqrt{\theta(1-\theta)}}{\theta r(\mu)+(1-\theta)r(L)}
 \leq\frac{u-v}{2\sqrt{r(\mu)r(L)}}.
 \label{eq:two-point-optimization}
\end{equation}
The final inequality follows from the arithmetic--geometric mean inequality. The stated endpoint weights make this inequality, the variance bound, and the chord bound equalities.
\end{proof}

\subsection{The minimax bias--conditioning frontier}
The endpoint bound identifies the smallest bias compatible with each conditioning requirement. Let $\kappa=L/\mu$ denote the condition number of the original curvature interval. A proximal filter achieves every target $\chi<\kappa$ by adjusting its regularization parameter, while a constant filter achieves zero bias at $\chi=\kappa$.

\begin{theorem}[Exact minimax frontier]
\label{thm:frontier}
For every $\chi\in[1,\kappa]$,
\begin{equation}
 \inf_{q>0\text{ continuous}:\,\chi(q)\leq\chi}\B(q)
 =\front(\chi)=\frac{L-\mu\chi}{2\mu L\sqrt\chi}.
 \label{eq:frontier}
\end{equation}
For $1\leq\chi<\kappa$, the following filter attains the infimum:
\begin{equation}
 q_\lambda(h)=\frac1{h+\lambda},\qquad
 \lambda=\frac{\mu L(\chi-1)}{L-\mu\chi}.
 \label{eq:optimal-lambda}
\end{equation}
For $\chi=\kappa$, any positive constant filter attains zero bias.
\end{theorem}

\begin{proof}
Every admissible filter must satisfy the lower bound generated by its endpoint values. Set $t=r(L)/r(\mu)\leq\chi\leq\kappa$. Optimizing the weights of a two-point population yields
\begin{equation}
 \B(q)\geq\frac{q(\mu)-q(L)}{2\sqrt{r(\mu)r(L)}}
 =\frac{1/\mu-t/L}{2\sqrt t}\geq\front(\chi),
 \label{eq:universal-lower}
\end{equation}
The middle expression decreases with $t>0$, proving the lower bound.

The proximal filter attains this lower bound because its curvature response is affine in its filter value. Specifically, $r(h)=1-\lambda q_\lambda(h)$, so the chord condition holds with equality, and
\begin{equation}
 \chi_\lambda=\frac{L(\mu+\lambda)}{\mu(L+\lambda)}.
 \label{eq:prox-condition}
\end{equation}
Lemma~\ref{lem:endpoint} therefore gives
\begin{equation}
 \B(q_\lambda)=\frac{L-\mu}{2\sqrt{\mu L(\mu+\lambda)(L+\lambda)}}
 =\front(\chi_\lambda).
 \label{eq:prox-bias}
\end{equation}
The choice in \eqref{eq:optimal-lambda} gives $\chi_\lambda=\chi$, including $\chi=1$ at $\lambda=0$. A constant filter handles $\chi=\kappa$.
\end{proof}

A two-client quadratic population realizes the frontier at any heterogeneity scale. Let $\zeta\geq0$ and consider an attaining filter with $r(L)/r(\mu)=\chi$. Choose
\begin{equation}
 \begin{aligned}
 p_\mu&=\frac{\chi}{1+\chi},& p_L&=\frac1{1+\chi},\\
 f_\mu(x)&=\frac\mu2x^2+\frac\zeta{\sqrt\chi}x,\\
 f_L(x)&=\frac L2x^2-\zeta\sqrt\chi\,x.
 \end{aligned}
 \label{eq:witness}
\end{equation}
These objectives satisfy $x_*=0$ and $\zeta_*=\zeta$, and substitution into \eqref{eq:bias-identity} gives $x_q=-\zeta\front(\chi)$. The construction chooses the weights as part of the population; it does not establish attainment for every prescribed set of client weights.

\subsection{Finite local steps and server convergence}
The frontier applies to finite local gradient steps and yields a convergence guarantee for a relaxed server update. We consider clients that perform a fixed number of gradient evaluations between communication rounds. A quadratic penalty centered at the server model includes proximal local training \citep{li2020}; setting the penalty to zero recovers ordinary local gradient descent.

\begin{corollary}[Finite local updates]
\label{cor:local}
\leavevmode\par\noindent
Fix $K\geq1$, $\lambda\geq0$, and $0<\eta\leq1/(L+\lambda)$. Client $i$ starts at $y_0=x$ and performs
\begin{equation}
 y_{k+1}=y_k-\eta\bigl(\nabla f_i(y_k)+\lambda(y_k-x)\bigr).
 \label{eq:local-recurrence}
\end{equation}
It sends $x-y_K$ to the server. Its filter is
\begin{equation}
 q_{K,\eta,\lambda}(h)=\frac{1-(1-\eta(h+\lambda))^K}{h+\lambda}.
 \label{eq:local-filter}
\end{equation}
Define $\chi_K=Lq_{K,\eta,\lambda}(L)/(\mu q_{K,\eta,\lambda}(\mu))$. Then $\chi(q_{K,\eta,\lambda})=\chi_K\in[1,\kappa]$ and
\begin{equation}
 \B(q_{K,\eta,\lambda})=\front(\chi_K).
 \label{eq:local-frontier}
\end{equation}
\end{corollary}

The filter satisfies the endpoint certificate because $r$ is concave as a function of $q$. For $K=1$, $q$ is constant and the bias is zero. For $K\geq2$, set $t=1-\eta(h+\lambda)$ and $S(t)=\sum_{j=0}^{K-1}t^j$. Then $q=\eta S(t)$ and $r=1-t^K-\lambda q$. On $t\in[0,1)$, differentiation shows that $r$ is concave in $q$, while $q$ decreases and $r$ increases with $h$. Lemma~\ref{lem:endpoint} applies. Appendix~\ref{app:local} provides the derivative calculation and derives the filter from the local recurrence.

Ordinary local gradient descent can attain every interior point of the frontier with finitely many local steps. Set $\lambda=0$. At $\eta=1/L$, the endpoint ratio is $[1-(1-\mu/L)^K]^{-1}$, which tends to one as $K$ grows. For fixed $K$, the ratio tends to $\kappa$ as $\eta$ tends to zero. Continuity therefore yields every $\chi\in(1,\kappa)$ for some finite $K$ and stable $\eta$. Averaging exact client minimizers attains the boundary $\chi=1$.

The conditioning criterion also gives a standard contraction bound for a relaxed server update. For a fixed quadratic population, choose $\alpha=2/(m_q+M_q)$ in $x^+=x-\alpha D_q(x)$. The spectral bound on $I-\alpha A_q$ implies
\begin{equation}
 \begin{aligned}
 \norm{x_T-x_*}&\leq\rho_q^T\norm{x_0-x_q}+\zeta_*\B(q),\\
 \rho_q&=\frac{\chi(q)-1}{\chi(q)+1}.
 \end{aligned}
 \label{eq:round-bound}
\end{equation}
This inequality separates convergence to the surrogate minimizer from its displacement relative to $x_*$. It neither establishes minimax optimality of the finite-time sum nor describes the unit server step without relaxation. Comparing total computation also requires accounting for the local work used to implement each filter.

\section{Proximal bias beyond quadratics and with inexact solves}
\label{sec:nonquadratic}
Proximal local training provides a way to extend the exact bias bound beyond quadratic objectives. Its first-order conditions retain a linear relation after replacing each Hessian by an average along a segment. We use this relation to prove a sharp nonlinear bound and then quantify the extra error when clients stop before solving their subproblems exactly.

\subsection{Exact proximal maps for smooth strongly convex objectives}
The proximal model represents clients that penalize deviation from the current server model while optimizing their local loss. Assume each $f_i$ is twice continuously differentiable on $\R^d$ and satisfies
\begin{equation}
 \mu I\preceq\nabla^2 f_i(x)\preceq LI\quad\text{for all }i,x.
 \label{eq:nonquadratic-assumptions}
\end{equation}
Retain $F,x_*,g_i$, and $\zeta_*$ from Section~\ref{sec:setup}. For a finite penalty parameter $\lambda\geq0$, define
\begin{equation}
 \begin{aligned}
 y_i(x)&=\argmin_y\{f_i(y)+\tfrac\lambda2\norm{y-x}^2\},\\
 D_\lambda(x)&=x-\sum_i p_i y_i(x).
 \end{aligned}
 \label{eq:prox-map}
\end{equation}
The penalty encourages the client solution $y_i(x)$ to remain close to $x$. At $\lambda=0$, each client returns its own minimizer.

The average proximal map has a unique fixed point $x_\lambda$ for every finite $\lambda\geq0$. Each $y_i$ is $\lambda/(\mu+\lambda)$-Lipschitz, so its weighted average is a contraction; at $\lambda=0$, the average map is constant.

\begin{theorem}[Sharp nonlinear proximal bias]
\label{thm:nonlinear}
\leavevmode\par\noindent
Under \eqref{eq:nonquadratic-assumptions},
\begin{equation}
 \begin{aligned}
 \norm{x_\lambda-x_*}&\leq\zeta_* B_\lambda,\\
 B_\lambda&=\frac{L-\mu}
 {2\sqrt{\mu L(\mu+\lambda)(L+\lambda)}}.
 \end{aligned}
 \label{eq:nonlinear-bias}
\end{equation}
The quadratic population in \eqref{eq:witness}, with $\chi=\chi_\lambda$, attains equality within this smooth strongly convex class. Moreover, all eigenvalues of the Jacobian of $D_\lambda$ lie in
\begin{equation}
 [m_\lambda,M_\lambda]=
 \left[\frac\mu{\mu+\lambda},\frac L{L+\lambda}\right].
 \label{eq:nonlinear-conditioning}
\end{equation}
The relaxed update $x^+=x-\alpha D_\lambda(x)$ with $\alpha=2/(m_\lambda+M_\lambda)$ is therefore a contraction with factor $(\chi_\lambda-1)/(\chi_\lambda+1)$.
\end{theorem}

The proof reduces the nonlinear fixed-point equations to the quadratic bias identity. Let $e=x_\lambda-x_*$ and $y_i=y_i(x_\lambda)$. Define the secant Hessians by
\begin{equation}
 H_i^{\rm sec}=\int_0^1\nabla^2 f_i(x_*+t(y_i-x_*))\,dt
 \label{eq:secants}
\end{equation}
These matrices satisfy the same curvature bounds. The proximal first-order conditions and the aggregate fixed-point relation yield
\begin{equation}
 \begin{aligned}
 (H_i^{\rm sec}+\lambda I)(y_i-x_*)&=\lambda e-g_i,\\
 \sum_i p_i(y_i-x_*)&=e.
 \end{aligned}
 \label{eq:secant-equations}
\end{equation}
These are the quadratic bias equations for $q_\lambda(h)=1/(h+\lambda)$. The uniform bound from Theorem~\ref{thm:scalar} applies to the secant matrices even though they depend on the fixed point. Appendix~\ref{app:nonlinear} gives the full proof and the Jacobian calculation; neither step requires a Hessian-Lipschitz assumption.

\subsection{A residual bound for inexact local solves}
Clients usually terminate local optimization after finite work, leaving a nonzero first-order residual. The server may also stop before reaching a fixed point of the aggregate update. The next result separates these two errors from the bias that remains even after exact local solves and full server convergence.

\begin{theorem}[An inexact-solve certificate]
\label{thm:residual}
\leavevmode\par\noindent
Assume \eqref{eq:nonquadratic-assumptions}. Let $x$ be any server model, let $y_1,\ldots,y_n\in\R^d$ be arbitrary client outputs, and let $\lambda\geq0$ be finite. Define the local residuals $e_i$, their RMS magnitude $\delta$, and the aggregate residual $a$ by
\begin{equation}
 \begin{aligned}
 e_i&=\nabla f_i(y_i)+\lambda(y_i-x),\\
 \delta&=\left(\sum_i p_i\norm{e_i}^2\right)^{1/2},&
 a&=x-\sum_i p_i y_i.
 \end{aligned}
 \label{eq:residuals}
\end{equation}
Then
\begin{equation}
 \boxed{\ \norm{x-x_*}\leq\zeta_*B_\lambda+
 \frac\delta\mu+\frac{\mu+\lambda}{\mu}\norm a.\ }
 \label{eq:residual-certificate}
\end{equation}
Each coefficient is separately sharp over the stated class.
\end{theorem}

The residual terms enter through an exact decomposition of the error. Form the secant Hessians in \eqref{eq:secants} with the current client outputs $y_i$, and set $Q_i=(H_i^{\rm sec}+\lambda I)^{-1}$ and $A=\sum_i p_i H_i^{\rm sec}Q_i$. Rearranging the residual equations gives
\begin{equation}
 A(x-x_*)=a-\sum_i p_iQ_i g_i+\sum_i p_iQ_i e_i.
 \label{eq:residual-equation}
\end{equation}
The scalar reduction bounds $\norm{A^{-1}\sum_i p_iQ_i g_i}$ by $\zeta_*B_\lambda$. The bounds $\norm{A^{-1}}\leq(\mu+\lambda)/\mu$ and $\norm{Q_i}\leq1/(\mu+\lambda)$ control the remaining terms. Appendix~\ref{app:residual} proves the inequality and separate sharpness of its coefficients.

Using the certificate requires curvature and heterogeneity bounds in addition to measurable residuals. Clients can evaluate $e_i$, and the server can compute $a$, but $\zeta_*$ depends on gradients at the unknown optimizer. A valid upper bound on $\zeta_*$ must therefore come from additional information. The certificate does not itself estimate this quantity.

More accurate local solves reduce the residual term without eliminating the worst-case fixed-point bias. At an exact aggregate fixed point, both residual terms vanish, yet the quadratic equality construction still attains $\zeta_*B_\lambda$. Increasing $\lambda$ lowers this bias bound while moving the uniform condition number toward $L/\mu$.

\section{Numerical illustrations and limitations}
\label{sec:checks}
Small numerical examples make the equality constructions and residual bounds reproducible without large training experiments. We evaluate the explicit quadratic instances, test the matrix identities, and solve a nonquadratic proximal example. The proofs establish the bounds; these computations illustrate them and check the accompanying implementation.

\subsection{Equality examples and algebraic checks}
\begin{figure}[t]
 \centering
 \pdfximage width \linewidth {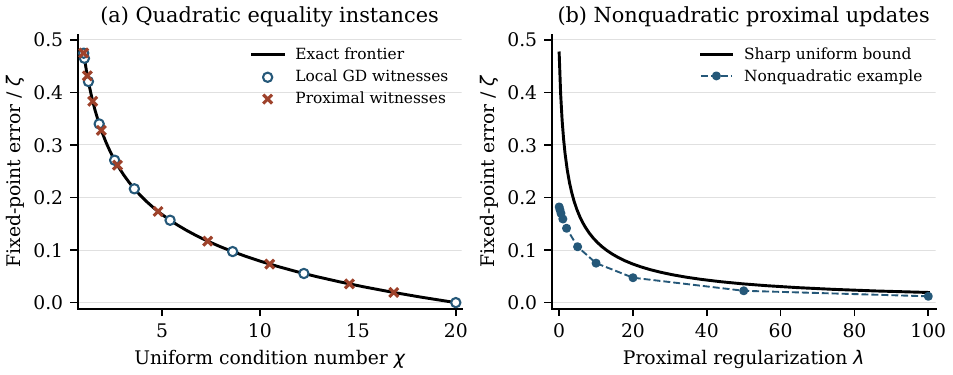}\pdfrefximage\pdflastximage
 \caption{Fixed-point bias for $\mu=1$, $L=20$, and unit heterogeneity. Left: the exact frontier and errors computed from the two-client equality instances for local gradient descent and proximal updates. Each point uses the population associated with its filter parameters. Right: the proximal upper bound and the computed error for the nonquadratic family in Appendix~\ref{app:repro}. The figure compares fixed-point errors; it does not measure training performance or total computation.}
 \label{fig:frontier}
\end{figure}

The numerical checks target equality on scalar instances, the matrix reduction, and the nonlinear residual bound. For the scalar checks, we solve the fixed-point equations for \eqref{eq:witness} at each filter's parameters and compare the error with \eqref{eq:frontier}. Figure~\ref{fig:frontier} shows these equality instances and a separate nonquadratic example. Appendix~\ref{app:repro} specifies all inputs, distributions, and solver tolerances.

The matrix checks test the reduction on clients whose Hessians have different eigenvectors. We generate $400$ populations, each with five $4\times4$ Hessians with independently sampled eigenbases. For fixed Hessians and weights, let $\B_{H,p}(q)$ denote the largest bias over mean-zero client gradients with RMS norm at most one. Appendix~\ref{app:identity} proves
\begin{equation}
 \B_{H,p}(q)^2=\lambda_{\max}\!\left(A_q^{-1}
 \left[\sum_i p_iQ_i^2-\overline Q^2\right]A_q^{-1}\right).
 \label{eq:instance-bias}
\end{equation}
We compute this value, reconstruct maximizing gradients, and compare the direct fixed-point error with the formula and uniform bound. All sampled populations satisfy the bound.

The three-point design example checks the optimality certificate on its full finite domain. We evaluate \eqref{eq:certificate} at every admissible curvature in Appendix~\ref{app:three} and recover the certified bias from the corresponding client population.

The numerical checks require only small CPU computations. The recorded run includes $74$ scalar equality cases, $400$ matrix populations, and $100$ nonlinear residual checks and takes approximately $3.24$ seconds, excluding dependency installation. The code, generated figure, and full results accompany the manuscript. These checks do not replace the proofs or establish validity for untested populations.

\subsection{Scope and limitations}
The lower bound applies to full deterministic aggregation with a fixed positive filter shared across clients. Persistent state, control variates, client-dependent or changing filters, and stochastic sampling fall outside this class. The finite-step result also requires quadratic objectives and $\eta\leq1/(L+\lambda)$; only the proximal result extends to the nonquadratic class. Because the supremum varies client weights and dimensions, a prescribed population can have smaller worst-case error.

The frontier records only curvature, optimum-gradient heterogeneity, and surrogate conditioning. Bounds on client-minimizer dispersion or Hessian similarity define different instance classes and may yield stronger guarantees. The analysis also omits local gradient-evaluation costs, so it does not select an optimal method under a finite computation budget. It makes no claim about generalization, privacy, fairness, or nonconvex learning.

\paragraph{Conclusion.}
The exact scalar characterization explains how curvature-dependent local updates convert client disagreement into fixed-point error. It yields a sharp bias--conditioning frontier attained by stable local gradient descent and proximal variants. The proximal bound also holds beyond quadratics and separates unavoidable worst-case bias from local-solve and aggregation residuals.

\clearpage
\subsection*{AI use statement}
In this work, an OpenAI Codex generative AI system contributed substantially to topic selection, literature discovery and synthesis, development and checking of theoretical models and mathematical claims, proof construction, research-method design, code and figure generation, interpretation of numerical checks, and manuscript drafting and revision. We did not use generative AI to generate empirical observations, collect or reformat datasets, translate the manuscript, or analyze human-subject data; those tasks are not part of this work. The numerical values reported in the paper were produced by the disclosed deterministic Python program rather than supplied by a language model. The author reviewed the AI-assisted material, checked the mathematical arguments against the stated assumptions, inspected the cited primary sources, and ran the accompanying verification code. No proof assistant or independent expert has certified the results. The author takes responsibility for the final content, including all text, claims, proofs, citations, code, figures, and other artifacts produced with AI assistance.

\bibliographystyle{plainnat}
\bibliography{references}
\clearpage
\section*{Checklist}

\begin{enumerate}

  \item For all models and algorithms presented, check if you include:
  \begin{enumerate}
    \item A clear description of the mathematical setting, assumptions, algorithm, and/or model. [Yes/No/Not Applicable]

    \textbf{Answer: Yes.} Sections~\ref{sec:setup}--\ref{sec:nonquadratic} define the objective class, heterogeneity budget, spectral updates, and local procedures.

    \item An analysis of the properties and complexity (time, space, sample size) of any algorithm. [Yes/No/Not Applicable]

    \textbf{Answer: Yes.} Corollary~\ref{cor:local} records the number of local gradient evaluations and Equation~\eqref{eq:round-bound} gives the server-round contraction; Section~\ref{sec:checks} states the limits of this cost model.

    \item (Optional) Anonymized source code, with specification of all dependencies, including external libraries. [Yes/No/Not Applicable]

    \textbf{Answer: Yes.} The anonymized supplementary archive contains the reproduction script, pinned dependencies, generated results, and an MIT license.
  \end{enumerate}

  \item For any theoretical claim, check if you include:
  \begin{enumerate}
    \item Statements of the full set of assumptions of all theoretical results. [Yes/No/Not Applicable]

    \textbf{Answer: Yes.} Each theorem states its assumptions, and Sections~\ref{sec:setup} and~\ref{sec:nonquadratic} define the common quadratic and nonquadratic settings.

    \item Complete proofs of all theoretical results. [Yes/No/Not Applicable]

    \textbf{Answer: Yes.} Proofs appear with the results or in Appendices~\ref{app:identity}--\ref{app:residual}.

    \item Clear explanations of any assumptions. [Yes/No/Not Applicable]

    \textbf{Answer: Yes.} The motivating paragraphs in Sections~\ref{sec:setup} and~\ref{sec:nonquadratic} connect the assumptions to local training, and Section~\ref{sec:checks} explains their practical limits.
  \end{enumerate}

  \item For all figures and tables that present empirical results, check if you include:
  \begin{enumerate}
    \item The code, data, and instructions needed to reproduce the main experimental results (either in the supplemental material or as a URL). [Yes/No/Not Applicable]

    \textbf{Answer: Yes.} The supplementary archive includes all code, recorded outputs, and instructions; the checks use no external dataset.

    \item All the training details (e.g., data splits, hyperparameters, how they were chosen). [Yes/No/Not Applicable]

    \textbf{Answer: Not Applicable.} The paper reports deterministic algebraic checks and a fixed-seed synthetic test rather than model training or evaluation on a dataset. Appendix~\ref{app:repro} specifies all numerical inputs and tolerances.

    \item A clear definition of the specific measure or statistics and error bars (e.g., with respect to the random seed after running experiments multiple times). [Yes/No/Not Applicable]

    \textbf{Answer: Not Applicable.} The numerical results verify identities and inequalities and do not estimate a population statistic. The text labels sampled maxima as checks rather than confidence bounds.

    \item A description of the computing infrastructure used. (e.g., type of GPUs, internal cluster, or cloud provider). [Yes/No/Not Applicable]

    \textbf{Answer: Yes.} Appendix~\ref{app:repro} and the recorded results identify the Python environment, CPU execution, processor information, and wall-clock time; no GPU or cluster was used.
  \end{enumerate}

  \item If you are using existing assets (e.g., code, data, models) or curating/releasing new assets, check if you include:
  \begin{enumerate}
    \item Citations of the creator If your work uses existing assets. [Yes/No/Not Applicable]

    \textbf{Answer: Yes.} The paper cites the prior methods and results it uses. The supplementary documentation identifies the numerical dependencies and the official AISTATS template.

    \item The license information of the assets, if applicable. [Yes/No/Not Applicable]

    \textbf{Answer: Yes.} The supplementary documentation records the dependency licenses; the dependencies themselves are not redistributed.

    \item New assets either in the supplemental material or as a URL, if applicable. [Yes/No/Not Applicable]

    \textbf{Answer: Yes.} The anonymized supplementary archive contains the new reproduction code, generated data, and figures.

    \item Information about consent from data providers/curators. [Yes/No/Not Applicable]

    \textbf{Answer: Not Applicable.} The work does not use data from external providers or curators.

    \item Discussion of sensible content if applicable, e.g., personally identifiable information or offensive content. [Yes/No/Not Applicable]

    \textbf{Answer: Not Applicable.} The work uses only mathematical constructions and synthetic numerical checks and contains no personal or offensive content.
  \end{enumerate}

  \item If you used crowdsourcing or conducted research with human subjects, check if you include:
  \begin{enumerate}
    \item The full text of instructions given to participants and screenshots. [Yes/No/Not Applicable]

    \textbf{Answer: Not Applicable.} The research does not use crowdsourcing or human participants.

    \item Descriptions of potential participant risks, with links to Institutional Review Board (IRB) approvals if applicable. [Yes/No/Not Applicable]

    \textbf{Answer: Not Applicable.} The research does not involve human participants.

    \item The estimated hourly wage paid to participants and the total amount spent on participant compensation. [Yes/No/Not Applicable]

    \textbf{Answer: Not Applicable.} The research does not involve participants or compensation.
  \end{enumerate}

\end{enumerate}

\clearpage
\appendix
\onecolumn
\aistatstitle{The Exact Bias--Conditioning Frontier of Local Updates: Supplementary Materials}
\section{Bias identities and fixed-population sharpness}
\label{app:identity}
The scalar reduction and matrix checks rely on exact operator expressions for the optimizer shift. We derive the bias identity, compute the largest shift for fixed Hessians and weights, and construct gradients that attain it. We also justify the dual norm identity used in the main proof.

The quadratic gradient formula gives the bias identity directly. Since $\nabla f_i(x)=H_i(x-x_*)+g_i$, substituting into $D_q(x_q)=0$ yields $A_q(x_q-x_*)+\sum_i p_iQ_i g_i=0$. This proves \eqref{eq:bias-identity}. Positivity and continuity of $q$ imply $m_q>0$ and $A_q\succeq m_q I$, so the inverse exists.

For fixed Hessians and weights, the largest bias is the norm of a linear map on mean-zero client gradients. Define $\mathcal G=\{(g_i):\sum_i p_i g_i=0\}$ with inner product $\langle g,h\rangle_p=\sum_i p_i g_i^\top h_i$ and induced norm $\norm g_p$. Let $T:\mathcal G\to\R^d$ act as $Tg=\sum_i p_iQ_i g_i$. Its adjoint is
\begin{equation}
 (T^*v)_i=(Q_i-\overline Q)v.
 \label{eq:adjoint}
\end{equation}
The right side has weighted mean zero, and subtracting $\overline Qv$ does not change its inner product with $g\in\mathcal G$. Consequently,
\begin{equation}
 TT^*=\sum_i p_i(Q_i-\overline Q)^2
 =\sum_i p_iQ_i^2-\overline Q^2=:V.
 \label{eq:operator-covariance}
\end{equation}
The cross terms simplify because $\sum_i p_iQ_i\overline Q=\sum_i p_i\overline Q Q_i=\overline Q^2$; this step requires no pairwise commutativity. Therefore
\begin{equation}
 \sup_{g\in\mathcal G:\,\norm g_p\leq1}\norm{A_q^{-1}Tg}^2
 =\lambda_{\max}(A_q^{-1}VA_q^{-1}),
 \label{eq:operator-norm}
\end{equation}
which proves \eqref{eq:instance-bias}.

An eigenvector construction attains this operator norm. Let $u$ be a unit top eigenvector of $A_q^{-1}VA_q^{-1}$, and write its eigenvalue as $\sigma^2>0$. Set
\begin{equation}
 g_i=\frac{(Q_i-\overline Q)A_q^{-1}u}{\sigma}.
 \label{eq:optimal-gradients}
\end{equation}
These gradients satisfy $\sum_i p_i g_i=0$, $\sum_i p_i\norm{g_i}^2=1$, and $A_q^{-1}Tg=\sigma u$. The objectives $f_i(x)=x^\top H_ix/2+g_i^\top x$ realize them with $x_*=0$. If $\sigma=0$, every admissible gradient collection gives zero shift.

The dual norm identity follows by changing variables through the symmetric invertible matrix $A_q$. For any $b\in\R^d$,
\begin{equation}
 \norm{A_q^{-1}b}
 =\sup_{u\ne0}\frac{|u^\top A_q^{-1}b|}{\norm u}
 =\sup_{v\ne0}\frac{|v^\top b|}{\norm{A_qv}},
 \label{eq:dual-identity}
\end{equation}
where the second equality uses $u=A_qv$ and symmetry of $A_q$. This is the identity applied in Theorem~\ref{thm:scalar}.

\section{Optimal-design representation, certificates, and support size}
\label{app:design}
The optimal-design representation turns the search for a worst-case population into a determinant maximization with a global certificate. We prove the change of measure, the optimality condition, and the three-point support bound in Proposition~\ref{prop:design}. The certificate specializes classical $D$-optimal design \citep{kiefer1960}; Theorem~\ref{thm:scalar} connects it to local-update bias.

Reweighting by $r$ converts the scalar variance ratio into an information-matrix determinant. Let $s=\E_Pr>0$ and define $d\nu=r\,dP/s$. Then
\begin{equation}
 M(\nu)=\frac1s
 \begin{pmatrix}1&\E_Pq\\\E_Pq&\E_Pq^2\end{pmatrix}.
 \label{eq:moment-matrix}
\end{equation}
Taking determinants proves \eqref{eq:tilt}. Conversely, for any measurable set $B$, define $P(B)=\int_B r^{-1}d\nu/\int r^{-1}d\nu$. Positivity and continuity of $r$ on the compact interval make the denominator finite and positive. This formula inverts the transformation and proves \eqref{eq:design}. Compactness of the space of probability measures and continuity of the matrix entries ensure that a maximizing information matrix exists.

The directional derivatives of the log determinant yield the necessary optimality condition. If $q$ is nonconstant, two vectors $z(h_1),z(h_2)$ are linearly independent, so a positive-definite information matrix exists. Any determinant maximizer $M_*$ is therefore positive definite and also maximizes the log determinant. For any $h$, the derivative at $\varepsilon=0$ of
\begin{equation}
 \log\det\bigl((1-\varepsilon)M_*+\varepsilon z(h)z(h)^\top\bigr)
 \label{eq:design-derivative-path}
\end{equation}
equals $z(h)^\top M_*^{-1}z(h)-2$. Optimality requires this derivative to be nonpositive, proving necessity of \eqref{eq:certificate}.

Concavity of the log determinant makes the same condition sufficient. Suppose $M_*$ satisfies \eqref{eq:certificate}, and let $M$ be any positive-definite feasible matrix induced by a measure $\nu$. Then
\begin{align}
 \log\det M-\log\det M_*
 &\leq\tr(M_*^{-1}(M-M_*))\\
 &=\int z(h)^\top M_*^{-1}z(h)\,d\nu(h)-2\leq0.
 \label{eq:design-sufficiency}
\end{align}
Singular feasible matrices have determinant zero and cannot improve the positive optimum. Thus \eqref{eq:certificate} characterizes optimality.

The certificate also restricts optimal information matrices to a two-dimensional affine space. Its left side has mean $2$ under an optimal measure, so equality holds almost everywhere under that measure. Hence $M_*$ lies in the convex hull of the matrices $z(h)z(h)^\top$ that belong to
\begin{equation}
 \{N\in\mathbb S^2:\tr(M_*^{-1}N)=2\}.
 \label{eq:support-plane}
\end{equation}
Here $\mathbb S^2$ is the three-dimensional space of symmetric $2\times2$ matrices. The displayed hyperplane has dimension two, so Carath\'eodory's theorem represents $M_*$ by at most three matrices on it. Applying the inverse change of measure gives client weights on the same support. The gradient construction in Theorem~\ref{thm:scalar} then attains the bias. For constant $q$, the bias is zero and one client suffices.

\paragraph{Restricted spectral sets.}
The same argument applies when the admissible curvature values form any nonempty compact set $S\subset(0,\infty)$. Assume $q$ is continuous and positive on $S$ and restrict each client spectrum to $S$. All maxima and certificates then range over $S$. The interval frontier in Theorem~\ref{thm:frontier} concerns the particular choice $S=[\mu,L]$.

\paragraph{Approximate certificates.}
A uniform approximate certificate bounds the gap between a candidate design and the optimum. Define the sensitivity of a positive-definite feasible matrix $M$ at $h$ as $z(h)^\top M^{-1}z(h)$. Let $\varepsilon\geq0$, and suppose this quantity is at most $2(1+\varepsilon)$ throughout $S$. For any feasible $M_*$,
\begin{equation}
 \sqrt{\det M_*}\leq(1+\varepsilon)\sqrt{\det M}.
 \label{eq:approx-certificate}
\end{equation}
To prove the bound, set $N=M^{-1/2}M_*M^{-1/2}$. The sensitivity assumption gives $\tr N\leq2(1+\varepsilon)$, and the arithmetic--geometric mean inequality gives $\sqrt{\det N}\leq\tr N/2$. On a finite spectral set, one can verify the certificate at every point. On a continuous interval, checking a finite grid alone does not establish the required uniform inequality.

\section{A three-client finite-spectrum example}
\label{app:three}
\leavevmode\par\noindent
Three clients can be necessary to attain the worst-case bias when the admissible spectral set is an arbitrary compact set. We demonstrate this with a three-point set and an algebraic optimality certificate. The example concerns a general filter, rather than the gradient-descent and proximal families covered by the endpoint lemma.

The construction chooses filter values that make all three support points active in the design problem. Let $S=\{1,5,10\}$, with $q(1)=1$, $q(5)=1/10$, and $q(10)=10$. Write $z_i=z(h_i)$ and define $C_{ij}=\det[z_i,z_j]^2$. Direct calculation gives
\begin{equation}
 C=\begin{pmatrix}
 0&81/50&81/100\\
 81/50&0&9801/5000\\
 81/100&9801/5000&0
 \end{pmatrix}.
 \label{eq:three-C}
\end{equation}
For design weights $w$, $\det M(w)=\tfrac12 w^\top Cw$. Choose $w=C^{-1}{\bf1}/({\bf1}^\top C^{-1}{\bf1})$, where $\bf1$ is the three-dimensional vector of ones. Every coordinate is positive, and numerically
\begin{equation}
 w\simeq(0.18315152,\ 0.44626546,\ 0.37058302).
 \label{eq:three-weights}
\end{equation}

These weights satisfy the optimality certificate at all three admissible curvatures. Write $Cw=c{\bf1}$, so $\det M=c/2$. Differentiating the determinant with respect to each design weight gives $z_i^\top M^{-1}z_i=(Cw)_i/\det M=2$. Proposition~\ref{prop:design} therefore proves global optimality on $S$.

The certified three-point optimum strictly exceeds every two-point value. Its bias is approximately $0.71523503$, whereas any two-point design has determinant at most $\max_{i<j}C_{ij}/4$ and bias at most $0.70003571$. Convert the design to client weights using $p_i=(w_i/r(h_i))/\sum_jw_j/r(h_j)$, then choose the centered, variance-normalized gradients from Theorem~\ref{thm:scalar}. The reproduction script implements the rational inputs and these transformations. The algebraic construction specifies the certificate; the displayed decimals summarize its numerical values.

\section{Stable finite local updates attain the frontier}
\label{app:local}
Finite local gradient steps attain the frontier because their spectral response satisfies the endpoint lemma. We derive that response from the client recurrence, prove its monotonicity and concavity properties, and then establish Corollary~\ref{cor:local}. The final paragraphs describe its small-step and many-step limits.

The local recurrence expresses the client displacement as a matrix geometric series. At client $i$, define $s_k=y_k-x$ and $G_i=\nabla f_i(x)$. For a quadratic objective,
\begin{equation}
 s_{k+1}=(I-\eta(H_i+\lambda I))s_k-\eta G_i,
 \qquad s_0=0.
 \label{eq:local-recurrence-proof}
\end{equation}
Iterating this relation gives
\begin{equation}
 x-y_K=\eta\sum_{j=0}^{K-1}(I-\eta(H_i+\lambda I))^jG_i
 =q_{K,\eta,\lambda}(H_i)G_i.
 \label{eq:filter-derivation}
\end{equation}
This recovers the standard spectral representation of local updates \citep{charles2021}.

The filter decreases with curvature, while the surrogate curvature response increases. For $K=1$, $q(h)=\eta$ and $r(h)=\eta h$, giving $\chi_K=L/\mu$ and zero bias. For $K\geq2$, set $t=1-\eta(h+\lambda)$ and $S(t)=\sum_{j=0}^{K-1}t^j$. The step-size assumption gives $0\leq t<1$, and $S'(t)>0$ implies that $q=\eta S(t)$ decreases strictly with $h$. Also,
\begin{equation}
 r(h)=\frac h{h+\lambda}\bigl(1-(1-\eta(h+\lambda))^K\bigr)
 =1-t^K-\lambda q.
 \label{eq:r-local}
\end{equation}
In the first expression, both factors are nonnegative and nondecreasing, the first is positive, and the second increases strictly. Thus $r$ increases strictly with $h$, with $m_q=r(\mu)$ and $M_q=r(L)$.

Concavity of $r$ as a function of $q$ establishes the chord condition. Differentiating on the interior gives
\begin{align}
 \frac{dr}{dq}&=-\frac{Kt^{K-1}}{\eta S'(t)}-\lambda,\\
 \frac{d^2r}{dq^2}
 &=-\frac{Kt^{K-2}\bigl((K-1)S'(t)-tS''(t)\bigr)}{\eta^2 S'(t)^3}\leq0,
 \label{eq:concavity-derivative}
\end{align}
because
\begin{equation}
 (K-1)S'(t)-tS''(t)
 =\sum_{j=1}^{K-1}j(K-j)t^{j-1}>0.
 \label{eq:positive-polynomial}
\end{equation}
Continuity extends concavity to $t=0$. Consequently $r(h)\geq\ell(q(h))$, and Lemma~\ref{lem:endpoint} gives the exact bias.

Rewriting the endpoint bias in terms of $\chi_K$ proves the frontier identity. Specifically,
\begin{equation}
 \frac{q(\mu)-q(L)}{2\sqrt{\mu Lq(\mu)q(L)}}
 =\frac{L-\mu\chi_K}{2\mu L\sqrt{\chi_K}}.
 \label{eq:local-simplification}
\end{equation}
Monotonicity of $r$ gives $\chi_K\geq1$, and monotonicity of $q$ gives $\chi_K\leq L/\mu$. This completes the proof of Corollary~\ref{cor:local}.

\paragraph{Small-step behavior.}
The exact formula also determines the leading bias term as the local step size tends to zero. For fixed $K\geq2$, $\lambda=0$, and $\eta\downarrow0$,
\begin{equation}
 \B(q_{K,\eta,0})=
 \frac{\eta(K-1)(L-\mu)}{4\sqrt{\mu L}}+O(\eta^2).
 \label{eq:small-step}
\end{equation}
The expansion $q(h)=K\eta-K(K-1)\eta^2h/2+O(\eta^3)$ holds uniformly on the fixed interval. Substituting it into \eqref{eq:endpoint-bias} proves the formula. This expansion concerns the worst case under the RMS-gradient budget, whereas an instance-specific vector expansion describes a fixed population.

\paragraph{Many-step limit.}
Increasing the number of stable local steps approaches the exact proximal filter. For fixed $\eta>0$ satisfying the corollary's step-size bound and fixed $\lambda\geq0$, the filter converges uniformly to $1/(h+\lambda)$ as $K\to\infty$. At $\lambda=0$, this limit averages client minimizers and has worst-case bias $(L-\mu)/(2\mu L)$.

\paragraph{Attaining interior condition numbers.}
Finite unregularized local gradient descent can attain any target $\chi\in(1,L/\mu)$. Choose $K$ so that $[1-(1-\mu/L)^K]^{-1}<\chi$. As a function of $\eta\in(0,1/L]$, the endpoint ratio is continuous, tends to $L/\mu$ at zero, and is below $\chi$ at $1/L$. The intermediate value theorem yields a positive step size with ratio $\chi$. This argument establishes existence, without minimizing the required local work.

\section{Nonquadratic proximal bias and contraction}
\label{app:nonlinear}
Theorem~\ref{thm:nonlinear} follows from three properties of proximal maps: contraction ensures a unique aggregate fixed point, secant Hessians recover the quadratic bias identity, and Jacobian bounds control the relaxed server update. We prove these properties under \eqref{eq:nonquadratic-assumptions}.

Strong convexity makes the proximal map well defined and contractive. Every local subproblem is $(\mu+\lambda)$-strongly convex and coercive and therefore has a unique minimizer. For $\lambda>0$, the implicit-function theorem applied to $\nabla f_i(y_i(x))+\lambda(y_i(x)-x)=0$ gives
\begin{equation}
 \nabla y_i(x)=\lambda\bigl(\nabla^2f_i(y_i(x))+\lambda I\bigr)^{-1}.
 \label{eq:prox-jacobian}
\end{equation}
Its norm is at most $\lambda/(\mu+\lambda)<1$. Averaging preserves this bound, so Banach's fixed-point theorem gives a unique $x_\lambda$. At $\lambda=0$, the average map is constant and has one fixed point.

The secant Hessians express the nonlinear fixed-point equations as an admissible quadratic system. Set $y_i=y_i(x_\lambda)$, let $H_i=H_i^{\rm sec}$ from \eqref{eq:secants}, and define $Q_i=(H_i+\lambda I)^{-1}$. The fundamental theorem of calculus gives $\nabla f_i(y_i)=g_i+H_i(y_i-x_*)$. Substitute this identity into the proximal first-order conditions, multiply by $Q_i$, and average. The fixed-point relation yields
\begin{equation}
 \left(I-\lambda\sum_i p_iQ_i\right)(x_\lambda-x_*)
 =-\sum_i p_iQ_i g_i.
 \label{eq:nonlinear-bias-identity}
\end{equation}
The left matrix equals $A=\sum_i p_iH_iQ_i$ because $I-\lambda Q_i=H_iQ_i$. Each $H_i$ satisfies the original curvature bounds, and the $g_i$ retain weighted mean zero and RMS norm $\zeta_*$. Thus the quadratic bound \eqref{eq:prox-bias} applies and proves \eqref{eq:nonlinear-bias}. Its uniformity over all admissible matrices permits the secant Hessians to depend on the fixed point.

The quadratic equality examples prove sharpness over the larger smooth function class. The objectives in \eqref{eq:witness} satisfy \eqref{eq:nonquadratic-assumptions} and attain $\zeta_*B_\lambda$ for $\chi=\chi_\lambda$. No smaller constant can therefore hold throughout that class.

The Jacobian of the aggregate direction determines the server contraction rate. For $\lambda>0$, \eqref{eq:prox-jacobian} implies
\begin{equation}
 \nabla D_\lambda(x)=\sum_i p_i\nabla^2 f_i(y_i(x))
 \bigl(\nabla^2 f_i(y_i(x))+\lambda I\bigr)^{-1}.
 \label{eq:aggregate-jacobian}
\end{equation}
The scalar function $h\mapsto h/(h+\lambda)$ increases, so the Jacobian eigenvalues lie in \eqref{eq:nonlinear-conditioning}. At $\lambda=0$, the Jacobian is $I$, as the same interval predicts. Set $\alpha=2/(m_\lambda+M_\lambda)$ and $\rho=(M_\lambda-m_\lambda)/(M_\lambda+m_\lambda)$. The Jacobian of $x\mapsto x-\alpha D_\lambda(x)$ then has norm at most $\rho$. Integrating along a segment proves a global contraction, hence
\begin{equation}
 \norm{x_T-x_*}\leq\rho^T\norm{x_0-x_\lambda}+\zeta_*B_\lambda.
 \label{eq:nonlinear-rounds}
\end{equation}
The bound assumes exact client directions and the stated server relaxation.

\section{Residual certificate and separately sharp constants}
\label{app:residual}
The residual certificate separates intrinsic fixed-point bias from errors in the local solves and aggregate fixed-point condition. We first derive the decomposition in \eqref{eq:residual-equation}, then bound its three terms and give examples that attain each coefficient separately.

Secant Hessians linearize the residual equations for arbitrary client outputs. Let $x,y_i,e_i,a$ have the definitions in Theorem~\ref{thm:residual}. Define $H_i$ by integrating the Hessian between $x_*$ and $y_i$, and set $u=x-x_*$. Then
\begin{equation}
 e_i=g_i+(H_i+\lambda I)(y_i-x_*)-\lambda u,
 \qquad \sum_i p_i(y_i-x_*)=u-a.
 \label{eq:inexact-secants}
\end{equation}
Define $Q_i=(H_i+\lambda I)^{-1}$ and $A=\sum_i p_iH_iQ_i$. Multiplying the first equation by $Q_i$ and averaging gives \eqref{eq:residual-equation}.

The uniform quadratic bound controls the centered-gradient term, while spectral bounds control the two residual terms. The admissible secant matrices and gradients give $\norm{A^{-1}\sum_i p_iQ_i g_i}\leq\zeta_*B_\lambda$ by Theorem~\ref{thm:scalar}. The remaining terms satisfy
\begin{align}
 \norm{A^{-1}a}&\leq\frac{\mu+\lambda}{\mu}\norm a,\\
 \norm{A^{-1}\sum_i p_iQ_i e_i}
 &\leq\frac{\mu+\lambda}{\mu}\sum_i p_i\norm{Q_i e_i}
 \leq\frac1\mu\sum_i p_i\norm{e_i}
 \leq\frac\delta\mu.
 \label{eq:residual-bounds}
\end{align}
Applying the triangle inequality to \eqref{eq:residual-equation} proves \eqref{eq:residual-certificate}.

Each coefficient is necessary even when the other two terms vanish. For the bias coefficient, use \eqref{eq:witness} with exact proximal solves at the aggregate fixed point, so $a=0$ and $\delta=0$. For the coefficient of $\delta$, take one client with $f(y)=\mu y^2/2$ and choose $x=y\ne0$. Then $g=0$, $a=0$, and $\delta=\mu|x|$. For the coefficient of $\norm a$, use the same objective and the exact solve $y=\lambda x/(\mu+\lambda)$. Then $g=0$, $\delta=0$, and $a=\mu x/(\mu+\lambda)$. All three examples satisfy the original upper curvature bound $L>\mu$. They establish separate sharpness, without asserting simultaneous equality for arbitrary residual magnitudes.

\section{Reproducibility and numerical results}
\label{app:repro}
The numerical checks evaluate the explicit equality cases, matrix identities, and residual bounds used in the analysis. The script \texttt{code/reproduce.py} regenerates Figure~\ref{fig:frontier}, the JSON results, and the table below. It uses CPU double-precision operations with Python, NumPy, SciPy, and Matplotlib; \texttt{code/requirements.txt} records the package versions. No external data, model checkpoint, network access, or GPU is required.

\paragraph{Scalar equality instances.}
The scalar checks compare directly computed fixed points with the frontier, including the stability boundary. We set $\mu=1$, $L=20$, and $\zeta=1$. For local gradient descent, we use $\eta=0.8/L$ and $K\in\{1,2,3,5,8,12,20,40,80,160\}$. For proximal updates, we use $\lambda\in\{0,0.2,0.5,1,2,5,10,20,50,100\}$. At each parameter setting, we evaluate the endpoint filter values, construct the weights and gradients in \eqref{eq:witness}, and compute $x_q=-\sum_i p_iq(h_i)g_i/\sum_i p_ih_iq(h_i)$. We compare $|x_q|$ with $\front(\chi(q))$. A separate check covers every combination of $K\in\{1,2,3,5,20,100\}$, $\lambda\in\{0,1,10\}$, and $\eta(L+\lambda)\in\{0.1,0.8,1\}$. Assertions use absolute tolerances to include zero-bias cases.

\paragraph{Noncommuting matrices.}
Independently sampled eigenbases test the matrix identities without imposing common eigenvectors across clients. We use NumPy's \texttt{default\_rng(731)} and generate $400$ populations. For each population, we sample five weights from $\operatorname{Dirichlet}(1,1,1,1,1)$. Independently for each client, we sample a $4\times4$ standard-normal matrix, take its QR orthogonal factor $U_i$, and sample four eigenvalues uniformly from $[1,20]$. We form $H_i=U_i\operatorname{diag}(h_i)U_i^\top$ and $Q_i=U_i\operatorname{diag}(1/(h_i+2))U_i^\top$, compute \eqref{eq:instance-bias}, and reconstruct maximizing gradients through \eqref{eq:optimal-gradients}. The script checks their weighted mean and RMS norm, compares direct fixed-point computation with the operator formula, and verifies the upper bound $B_2$.

The reported matrix maximum summarizes only the sampled populations. It is a reproducible check of the formulas, rather than an estimate of expected algorithmic performance or a confidence bound for all populations. We therefore report the maximum ratio to the sharp bound without statistical error bars.

\paragraph{Nonquadratic example.}
The nonlinear example tests the proximal theorem on objectives with curvature that varies with the iterate. For the right panel of Figure~\ref{fig:frontier}, we use
\begin{equation}
 f_1(x)=\tfrac12x^2+c_1x+\log\cosh x,\qquad
 f_2(x)=9x^2+c_2x+2\log\cosh x.
 \label{eq:nonlinear-example}
\end{equation}
The $\log\cosh$ terms give Hessians in $[1,2]$ and $[18,20]$, respectively, preserving the common bounds $\mu=1$ and $L=20$. We set $(p_1,p_2)=(4/5,1/5)$ and $(c_1,c_2)=(1/2,-2)$, which give $x_*=0$ and $\zeta_*=1$.

Scalar root finding lets us check the bias and residual bounds on this nonlinear family. For each of the ten penalty values above, we solve the strictly increasing first-order equations and aggregate fixed-point equation with Brent's method. The script expands symmetric brackets until it finds a sign change and uses absolute tolerance $10^{-13}$ and relative tolerance $10^{-14}$. It checks the bias bound and residual certificate at the fixed point, at four additional deterministic query points, and after perturbing the local solutions by $(0.013,-0.021)$. These perturbations exercise the nonzero-residual terms. The quadratic constructions establish sharpness; this example checks the inequality on nonquadratic objectives.

\paragraph{Three-point design.}
The finite spectral example permits a complete numerical check of the design certificate. We use the inputs in Appendix~\ref{app:three}, solve the displayed linear system for $w$, and form $M$. The script checks positivity of the weights and verifies that all three sensitivities equal $2$. It then converts to client weights using $p_i\propto w_i/r(h_i)$ and checks that the resulting quadratic bias equals $\sqrt{\det M}$. Because the domain contains exactly three points, the check covers the entire design domain.

\begin{table}[ht]
\centering
\caption{Reproducible numerical checks. Errors are absolute floating-point discrepancies. The matrix ratio is a maximum over sampled instances, not a confidence bound.}
\label{tab:checks}
\begin{tabular}{lr}
\toprule
Quantity & Recorded value \\
\midrule
Scalar witness configurations & 74 \\
Maximum witness discrepancy & $1.11\times10^{-16}$ \\
Noncommuting matrix populations & 400 \\
Maximum matrix bias / sharp uniform bound & 0.904697 \\
Maximum matrix identity discrepancy & $5.27\times10^{-16}$ \\
Nonquadratic residual checks & 100 \\
Three-point optimum & 0.71523503 \\
Best two-point value on the same spectrum & 0.70003571 \\
Total script wall time (seconds) & 3.24 \\
\bottomrule
\end{tabular}
\end{table}

\paragraph{Runtime and environment.}
The reported runtime includes the numerical checks, root finding, and figure generation. It excludes dependency installation. The results file records Python and package versions, the operating-system family, and processor information, without user-identifying paths.

\paragraph{Reproduction commands.}
The source package contains everything needed to reproduce the calculations and compile the paper. From the project root, install \texttt{code/requirements.txt} and run \texttt{python code/reproduce.py}; the script raises an assertion error if a check fails. To build the paper locally, run pdfLaTeX on \texttt{main.tex}, run BibTeX, and run pdfLaTeX twice more. Overleaf can compile the included figures and generated table without running Python. Successful floating-point checks support reproducibility; the written proofs establish the theoretical results.

\clearpage
\section{Research transparency and broader impact}
\label{app:transparency}
Research transparency requires stating both the assistant's role and the limits of the validation performed. An OpenAI Codex language-model assistant contributed substantially to topic selection, literature search, proof development and checking, manuscript drafting and revision, reproduction code, and self-review. Its contribution extended beyond copyediting. The manuscript includes the mathematical arguments and executable checks, but no proof assistant or independent expert has certified the results. Human authors remain responsible for validating the claims, originality, and checklist before submission; a language model is not an author.

The results concern optimization error and have limited implications for the deployment of distributed learning systems. They may help compare local-update parameters, but they do not establish generalization, privacy, fairness, or security guarantees. The work uses no human subjects, personal data, or trained models. We identify no direct harmful application specific to these results beyond the general risks of deploying learning systems without appropriate validation. Improved surrogate conditioning alone does not imply a statistical or societal benefit.

\end{document}